\documentclass[sigconf,natbib=false]{acmart}
\usepackage[style=ACM-Reference-Format,backend=bibtex,sorting=none]{biblatex}
\usepackage[lofdepth,lotdepth]{subfig}

\setcopyright{none}
\copyrightyear{2021/2022}
\acmDOI{}
\acmISBN{}
\acmConference[Data Science Lab]{Data Science Lab}{2021/2022}{Uni Passau}

\begin{document}

\title{To GAN or Not To GAN: Segmentation Analysis on Mars DEM}

\author{Douglas D. Agbeve}
\email{agbeve01@ads.uni-passau.de}
\affiliation{\institution{University of Passau}}
  
\author{Aditya V. Handrale}
\email{handra01@ads.uni-passau.de}
\affiliation{\institution{University of Passau}}
  
\author{Salim Fares}
\email{fares01@ads.uni-passau.de}
\affiliation{\institution{University of Passau}}
  
\author{Seif E. Idani}
\email{idani01@ads.uni-passau.de }
\affiliation{\institution{University of Passau}}

\renewcommand{\shortauthors}{Team 5}

\begin{abstract}
\textit{
To better understand Martian Surface, which is needed to enable Rovers navigate Mars with ease, it is necessary to be able to determine the location of mounds. Detecting and studying these morphologies can also help us find evidence of extraterrestrial life, in this case, more specifically, water or signs of life conducive environments. Detection of mounds was done by manually mapping morphological parameters onto Digital Elevation Models.
This paper solves the problem by automatically detecting and or predicting mounds on Mars using Neural Network based Semantic Segmentation methodologies. This is done by using supervised semantic segmentation model and generative adversarial approach. A comparison of the approaches shows that adding extra artificially generated data did not improve the result.
}
\end{abstract}

\maketitle
\section{Introduction}
A picture, they say, is worth a thousand words. As a medium of communication,
pictures have been used since the dawn of \textit{Homo Sapiens}\cite{Yuval},
which makes understanding the information embed in them a major part of research
in the various fields of science ranging from Archaeology to the relatively
more recent Computer Science.\par
In Computer Vision, a vital process in retrieving information in images and
which form part of most systems for visual understanding is image
segmentation\cite{ForsythPonce}. It is the process of partitioning an image or a
video frame into multiple segments\cite{YingTan}. Segments are fundamentally a
set of pixels in a region that share a common property such as colour or
texture, enabling the identification and locating of boundaries of objects in an
image\cite{DassDevi}. Digital image segmentation is applied in various domains, 
including but not limited to, medical (locating cancerous tumors and measuring
tissue volumes)\cite{PhamXuPrince}, facial and fingerprint recognition,
self-driving vehicles (detecting pedestrian).\par There are, and continue to be
proposed, many algorithms for image segmentation. Currently, these approaches are
either segregating pixels based on intensity changes (i.e.~detecting discontinuity)
such as edge detecting algorithms, or partitioning an image into regions with
similar predefined criteria (i.e.~Similarity detection) examples include
thresholding approaches. The various algorithms can also be categorized based on
the method used in solving the problem; namely (1) Edge detection based
segmentation e.g.~histogram-based\cite{ZengLiMengYangLiu} and
gradient-based\cite{SaifHammadAlqubati}, (2) Thresholding approaches such
as those proposed by Abd Elaziz et al.\cite{ElazizBhattacharyyaLu}, and
Houssein et al.\cite{HousseinEmamAli}, (3) Region-based segmentation
techniques include region growing\cite{ChengWang, Jothiaruna}, region split and
merge\cite{Lachaize, LiuSclaroff}, (4) techniques based on Partial Differential
Equation are active contour\cite{KassWT}, C-V model\cite{WangWWF} etc., (5)
Clustering methods such as K-means clustering algorithm\cite{YanCaiGaoLuo} and
relatively more recent approach is the Artificial Neural Network (ANN) algorithms
which solves image segmentation problems with higher accuracy compared to other
approaches. Examples of Deep learning or ANN approach to image segmentation
include, but not limited to, Convolutional Network models such as VGG16,
GoogLeNet, Fast R-CNN, Faster R-CNN\cite{ren2016faster},
U-NET\cite{ronneberger2015unet}, V-NET\cite{milletari2016vnet} and
Mask-RCNN\cite{he2018mask}, Recurrent Neural Network models include
ReSeg\cite{visin2016reseg},Generative and Adversarial (GAN) Models e.g.
\@\cite{luc2016semantic, 8237868,hung2018adversarial}, a comprehensive list of
GAN-based techniques can be found in~\cite{ganzoo}.
\par
\begin{figure}[ht]
    \centering
    \subfloat[input image][Input Image]{
        \includegraphics[width=0.225\textwidth]{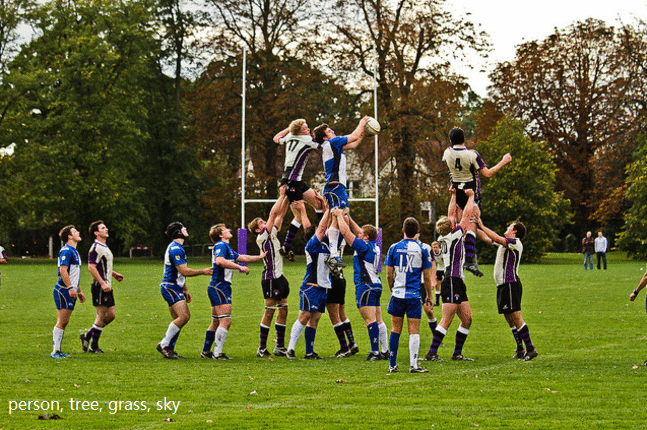}\label{fig:inputImage}
    }
    \subfloat[object detection][Object Detection]{
        \includegraphics[width=0.225\textwidth]{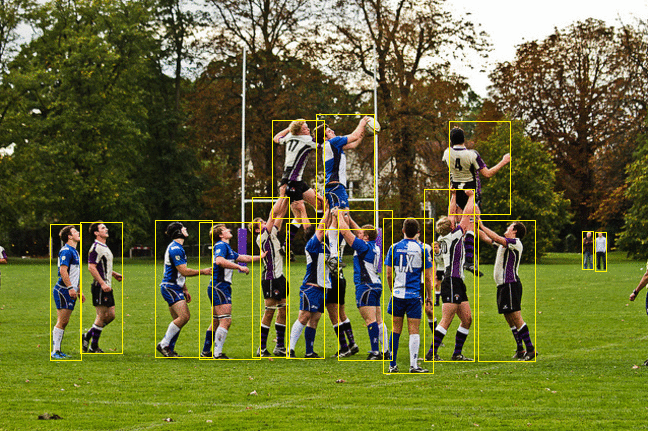}\label{fig:ObjDet}
    }
    \newline
    \subfloat[semantic image][Semantic Segmentation]{
        \includegraphics[width=0.225\textwidth]{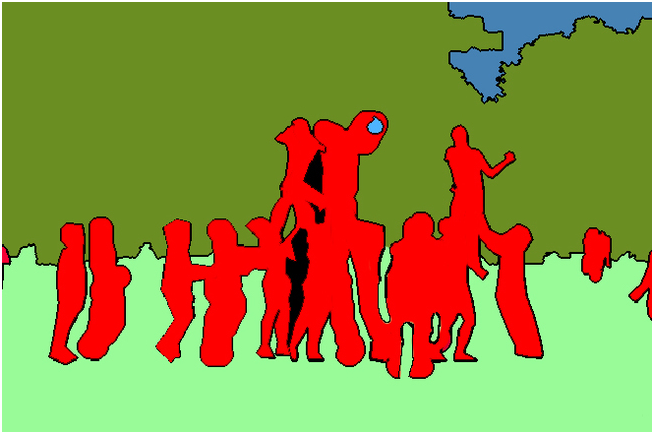}\label{fig:SSeg}
    }
    \subfloat[instance image][Instance Segmentation]{
        \includegraphics[width=0.225\textwidth]{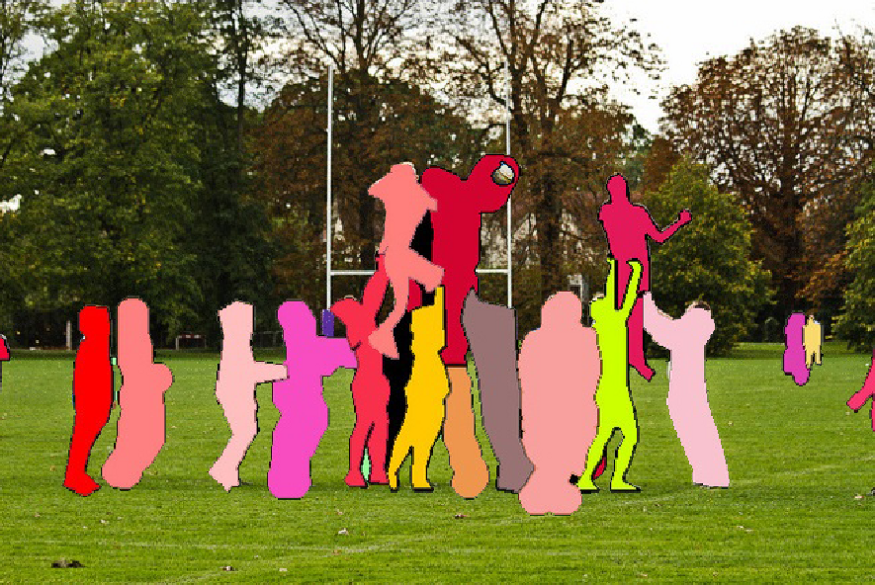}\label{fig:ISeg}
    }
    \newline
    \subfloat[panoramic image][Panoramic Segmentation]{
        \includegraphics[width=0.3\textwidth]{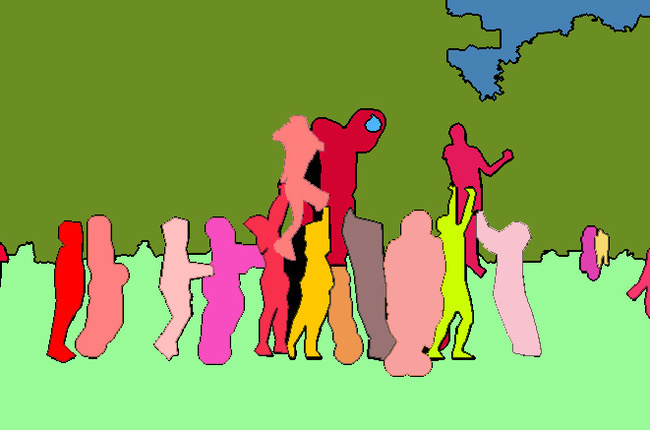}\label{fig:PSeg}
    }

    \caption{Variations of Segmentation~\cite{BigdataAILab}}
\end{figure}

The problem of image segmentation can be coined as that of classifying pixels
with semantic labels i.e.~semantic segmentation (fig.~\ref{fig:SSeg}) or
segregating pixels into individual objects in the image i.e.~instance
segmentation (fig~\ref{fig:ISeg})\cite{PaneruJeelani}. In semantic segmentation,
pixels are assigned to the same segment if they are of the same object type in
the image. This can be thought of as classification at pixel level. Instance
segmentation involves assigning all the pixels that belong to the same single
object to the same segment. It is a two-step process; extracting bounding
boxes around each instance of an object through object detection and then
classifying pixels that  corresponding to each instance in the bounding box.
This technique combines object detection (fig.~\ref{fig:ObjDet}) and
segmentation. A third, relatively new, formulation is to combine both instance
and semantic segmentation termed Panoramic Segmentation (fig.~\ref{fig:PSeg}).
It involves the detection and segmentation of all objects including background
and labelling different instances in an image.\par
Digital Elevation Models (DEMs) are essential in analyzing erosion and drainage,
hill-slope hydrology, studying groundwater flow, watersheds, and contaminant
transportation as they are important tools for parameterizing topography. DEM is
representation of planetary (earth, moon, mars etc.) terrain from elevation data
in 3-D image. There are two types; raster Geographic Information Systems (GIS)
layer representation and the vector-based triangular irregular network format.
DEMs are obtained using techniques such as Surveying, Stereo Photogrammetry,
Lidar, Radar, etc.. Flight and Train simulations, GIS and Satellite navigations
are some of the systems in which DEMs are used. Martian surface has an abundance
of geographical features such as volcanoes, layers and gullies, with phenomena
like volcanoes bringing out microbial entities that were protected from solar
radiation. Moreover, glaciers and mounds that have water underneath them might
contain evidence of life. Detecting and studying these morphologies can help us
find extraterrestrial life on mars. The detection of mounds formed from
phenomena such as volcanic eruption can be done automatically using ANN based
image segmentation methods.
Wuff-Jensen et al. proposed, to the best of our knowledge, the original
idea of using GANs to generate terrain from DEMs. Their architecture was based
on deep convolutional GAN (DCGAN)\cite{RMC}~--~a type of GAN made up of a
fractional-strided convolutions (generator) and a discriminator of strided
convolutions.\par
Spick et al.\cite{SpickCW} presented a method of generating height maps
from digital elevation of regions of earth. The authors approach was based on
Spatial GANs (SGAN)\cite{JetchevBV}. Spatial GANs remove fully connected layers
from DCGAN allowing outputs to be scaled to any size and mapping features onto
output as translation-invariant.\par
Bowles et al.\cite{bowles2018gan} investigated augmenting training data
using GAN-derived synthetic images. Progressive Growing of GANs
(PGGAN)\cite{KarrasAJ} network was used as a baseline architecture, and
demonstrated that this can improve results across two segmentation tasks
(1)~-~Computed Tomography (CT) images with manually delineated Cerebrospinal
Fluid (CSF) labels, (2)~-~Fluid-Attenuated Inversion Recovery (FLAIR) images
with manual White Matter Hyperintensity (WMH) segmentations.
\subsection{Research Questions}
The many algorithms available for solving image segmentation problem is an
indication that no single approach can solve it all accurately and in a more
efficient way. An affirmation of this assertion is the numerous methodologies
proposed in literature by Deep Learning researchers in recent decade.
Furthermore, the choice of algorithm and by extension the need for new
approaches, is principally influenced by data pertaining to the problem, its
representation and the type of segmentation problem.
In view of these, we formulated the following questions with regard to the data
at hand.
\begin{itemize}
    \item Will data augmentation have any effect on the accuracy of our selected
        approach (es) for application, and if yes, how much extra data relative
        to the training data?
    \item Which Deep learning technique(s) segments the data with better accuracy?
\end{itemize}

\section{Problem statement}
In this section, we formally present the problem of automatically detecting
mounds in Mars' Arabia Terra. Specifically, classification of pixels in the
digital elevation model of Mars in to segments of mounds or no-mounds using
using Artificial Neural Network based techniques. We sought to predict where
mounds given any DEM of Mars\par
More formally, given $n$ annotated images $\lbrace x_{1},x_{2},\ldots,x_{n}
\rbrace$ and each image, say $x_{i}$, has $m_{i}$ objects that are categorize
into $C$ classes with objects labelled as $y_{i}$:
\begin{equation}
    y^{i} = \lbrace (c^{i}_{1},v^{i}_{1}), (c^{i}_{2},v^{i}_{2}, \ldots,
    (c^{i}_{m_{i}},v^{i}_{m_{i}})) \rbrace,
\end{equation}
\hspace{2em} {$where$} $c^{i}_{m_{i}}\in C$ and $v^{i}_{m_{i}}$ is the object's
pixel mask.\\
The goal is to learn the values of some parameterized $(\theta)$ function $f$
such that:
\begin{equation}
    y^{i} = f(x_{i},\theta_{i})
\end{equation}
And be able to predict object (mound) locations $\hat{y}$ for all new (unseen)
$x_{i}$\\
A loss function to optimize the prediction would be:
\begin{equation}
    J(\theta) = \dfrac{1}{n}\sum_{i=1}^{n}l(\hat{y}^{i},x_{i},y^{i};\theta) +
    \alpha\lambda( \theta ),
\end{equation}
\hspace{2em} {$where$} $\alpha \in [0, \infty)$ controls the
relative contribution of the penalty norm, $\lambda$, in other words, it
controls the strength of the regularizer. No regularization when $\alpha = 0$,
with larger values indicating more regularization.\\
$\lambda(\theta) = \dfrac{1}{2}\|\theta \|_{2}^{2}$\hspace{1.0em}for $L_2$
regularization and \hspace{0.5em}
$\lambda(\theta) = \|\theta \|_{1}$\hspace{1.0em}for $L_1$ regularization.
\par
Due to the relatively small number of training samples, we proposed, as a
starting point of our implementation, inspired by the work done by Souly
\textit{et al.}\cite{8237868}, an architecture with a generator to provide extra
training samples and classifier as a discriminator in a Generative Adversarial
Network. We evaluate the performance of each approach against the following
metrics \textit{Pixels Accuracy} i.e.~ratio of properly classified pixels to
total number of pixels, \textit{Mean Pixel Accuracy (MPA)} is the average of the
ratio of properly classified pixels to the number of pixels in a class and
\textit{Mean Intersection over Union (mIoU)} is the ratio of the intersection
between the ground truth and the predicted segmentation map to their union,
averaged over all classes.
\par
The paper follows the following structure: \textit{section 3} discusses how the data was acquired
and the preprocessing techniques such as feature engineering that were used followed with, in \textit{section 4}, an explanation of the methods used to solve the problem at hand and finally in \textit{section 5}, evaluation of our methodology and results are discussed.

\section{Data Acquisition \& Pre-Processing}
\subsection{Data acquisition}
 In our paper, we are using data from HiRise (High Resolution Imaging Science Experiment) camera, onboard the Mars Reconnaissance Orbiter (MRO) spacecraft. We have the DEM (Digital elevation model) of the ‘Firsoff‘ crater, which is an impact crater on mars. Two high resolution images of a specific area on the ground are taken from different camera angles, and then these stereo images are combined together to obtain a DTM (Digital Terrain Model). The photos were taken at an altitude of 272 kilometers above ground. The scale of the image is very high in resolution at 0.27 meter per pixel. This data has been made publicly available by NASA/JPL/ University of Arizona. 
 We decided to utilize a DEM format, since it is a three dimensional digital representation of a terrain with X,Y and Z coordinates. The Z coordinate (Elevation data) will help us find natural morphologies on the surface like mounds, craters and channels. We can use DEMs to generate additional features by performing slope and hillshade analysis. The use of DEMs also allows us to build 3D models of the surface. Moreover the distance between the sample points (spatial resolution) and the vertical resolution are very high in the HiRise DEM.

\subsection{Data preprocessing}
The first step in preprocessing was to fill in the No Data values. A no data value is present in the dataset when there is no depth reading available at that coordinate. This usually represents things such as backgrounds and borders, but also occurs when data is not available due to technical difficulties etc. 

There are a few methods that can be used to treat the NoData values. We chose to perform automatic interpolation using the ‘fillnodata’ method from the python package ‘rasterio’\footnote{\url{https://rasterio.readthedocs.io/en/latest/api/rasterio.fill.html}}. Manual interpolation is also possible, but requires expertise in domain. Sometimes interpolation can also cause geometric patterns and artifacts in the image which end up adding noise to our dataset.

The rasterio interpolation we used has two tunable parameters, Max\_search\_distance and smoothing\_iterations. Max\_search\_distance corresponds to the maximum number of pixels to search in all directions to find values to interpolate from (using inverse distance weighting). The smoothing\_iterations refers to the number of 3x3 average filters (passes to run on interpolated pixels) are applied to smooth out artifacts. We used a value of 60 for the max\_search\_distance and a value of 0 for the smoothing iterations. Those configurations allowed us to interpolate the area of interest. We did not want to interpolate the entire image because that would add a lot of noise to the dataset.

After interpolating, the DEM file is now ready for the tiling process.
\begin{figure}[ht]
    \centering
    \subfloat[Interpolation][Interpolation]{
    \hspace*{-0.5cm}
        \includegraphics[width=0.45\textwidth]{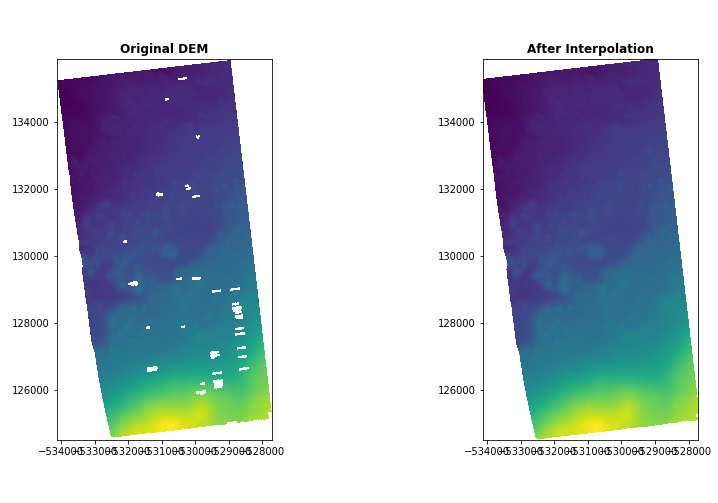}\label{fig:Interpolation}
    }
    \caption{The purpose of using interpolation is to fill the missing values in the original DEM, within the area of interest, as in the resulted image on the right.}
\end{figure}

The DEM image we are using has a resolution of 6418 x 11339, and is a geoTIFF file. Each pixel corresponds to the depth measurement at a specific area on the ground. The maximum depth value is -1595.222, maximum value is -3050.866 and the No Data value is represented by -32767. The DEM has 75\% valid values and a mean of -2594.830. A standard deviation value of 336.037 was observed.

\subsection{Feature engineering}
Adding channels or features to our data will help the segmentation neural network detect morphologies better. We can make use of additional features such as hillshades, slope and aspect analysis etc. We decided to generate hillshade and slope from our DEM file as additional features using gdal.DEMProcessing
\footnote{\url{https://gdal.org/python/osgeo.gdal-module.html\#DEMProcessing}}.
\begin{figure}[ht]
    \centering
    \subfloat[origin image][Origin Image]{
        \includegraphics[width=0.35\textwidth]{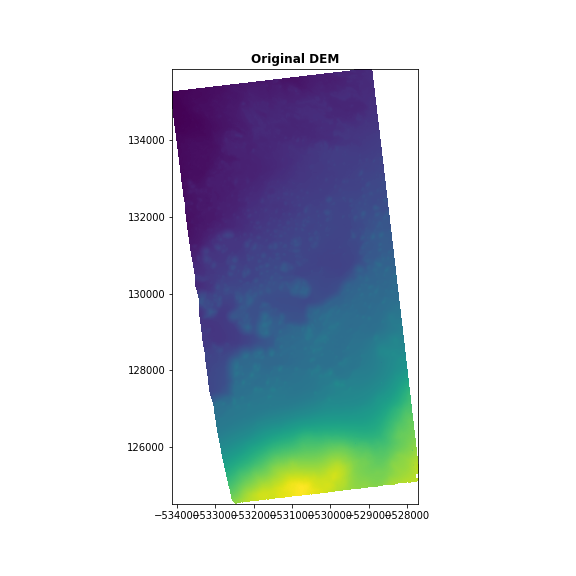}\label{fig:Origin Image}
    }
    \newline
    \subfloat[slope][Slope]{
    \hspace*{-1.5cm}
        \includegraphics[width=0.35\textwidth]{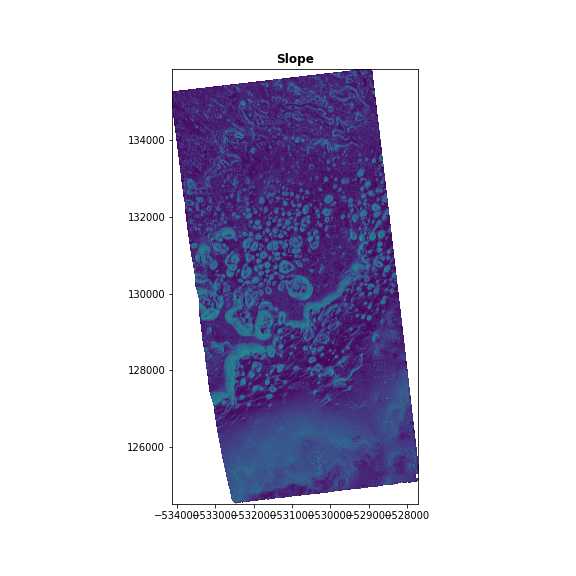}\label{fig:Slope}
    }
    \subfloat[hillshade][Hillshade]{
    \hspace*{-1.5cm}
        \includegraphics[width=0.35\textwidth]{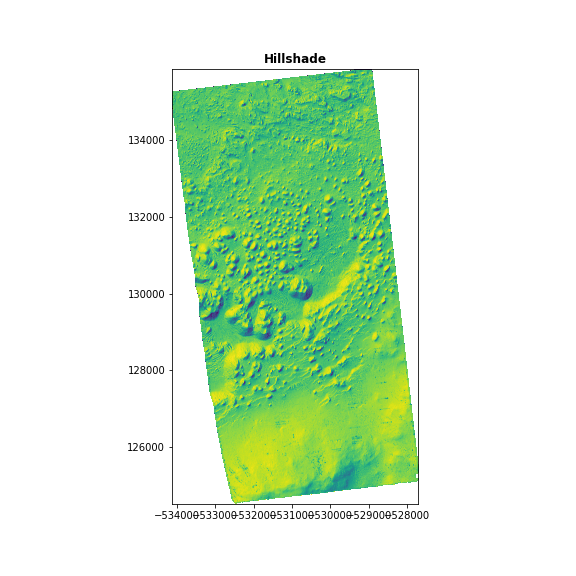}\label{fig:Hillshade}
    }
    \caption{Slope and Hillshade gave us a better visualization of the terrains, which is why we used them as additional channels for the segmentation model.}
\end{figure}

Since the dataset is very small and consists of only one image, we decided to tile the image into XDIV x YDIV parts. Tiling will allow us to have multiple training samples, which in turn will improve the performance of our neural network. All of the morphologies to be detected have a polynomial, amoeba-like shape. There arise a few problems when using tiling, ex. A tile border can cut through a morphology, giving it a linear edge where there is none. Since we will also be tiling the testing data set, this should not be a problem. We created multiple sets with different (XDIV x YDIV) values to have more diversity in our dataset.

(XDIV x YDIV) = (2 x 4), (4 x 8), (8 x 16), (16 x 32), (32 x 64)

Which gave us in total 2,728 tiles. We would split them into train, validation and test sets (We used some of the tiles for testing, since the actual test set does not have labels). 

To tile the DEM, firstly, find the top-left corner of the image which has the lowest x-coordinate and the highest y-coordinate using gdal.GetGeoTransform.

(xmin, ymax) = (gdal.GetGeoTransform()[0],gdal.GetGeoTransform()[3])

Then, get the size of the raster image which is the length of the coordination axis multiplied by the respected pixel size:

xlen = xres * gdal.RasterXSize; xres = gdal.GetGeoTransform()[1]

ylen = yres * gdal.RasterYSize; yres = gdal.GetGeoTransform()[5]

After that, define the number of divisions on each axis. (xdiv, ydiv) which will make the size of the tile as follows:

 xsize = xlen/xdiv
 
 ysize = ylen/ydiv

Next, define the strides (steps) of tiling as follows:

xsteps = [xmin + xsize * i for i in range(xdiv+1)]

ysteps = [ymax - ysize * i for i in range(ydiv+1)]

Finally, using gdal.Warp for each step in the steps vectors, passing these parameters:
\begin{itemize}
    \item The wanted name of the output.
    \item The input DEM.
    \item outputBounds=(xsteps[i], ysteps[i], xsteps[i+1], ysteps[i+1]
\end{itemize}
\begin{figure}[ht]
    \centering
    \subfloat[original dem][Original DEM]{
    \hspace*{-1.5 cm}
        \includegraphics[width=0.35\textwidth]{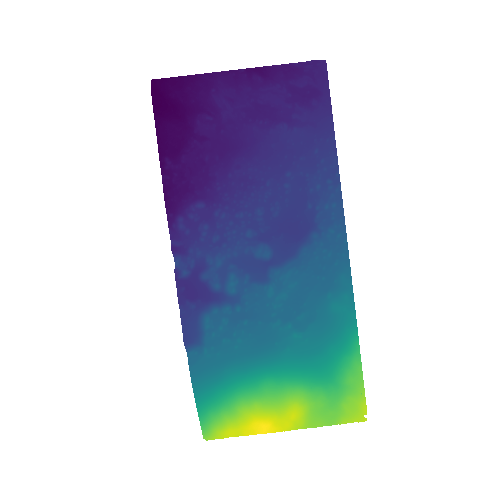}\label{fig:Original DEM}
    }
    \subfloat[tiles][Tiles]{
    \hspace*{-2.0cm}
        \includegraphics[width=0.35\textwidth]{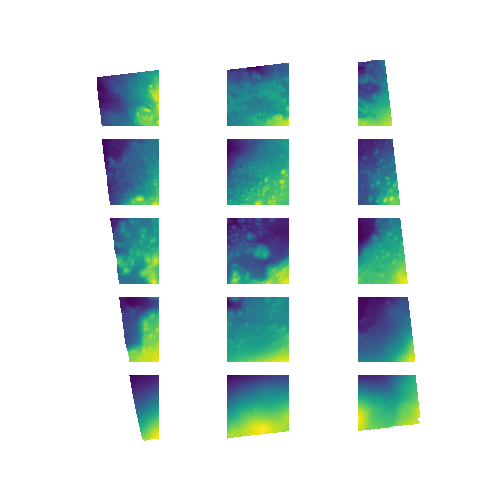}\label{fig:Tiles}
    }
    \caption{Splitting the original DEM into a set of tiles to. In this example the original DEM is split into 12 equal-sized tiles using the tiling algorithm explained in this paper, with xdiv = 3 (corresponds to the number of resulting columns) and ydiv = 4 (corresponds to the number of resulting rows)}
\end{figure}
\subsection{Annotation}
To annotate our dataset we need to extract the mounds from our DEM using the shape file that describe the mounds. 
We use the geopandas.read\_file()\footnote{\url{https://geopandas.org/en/stable/docs/reference/api/geopandas.read_file.html}} to read the shape file. We reproject the labels coordinate system to that of the original DEM which is stored in the ‘meta’ field of the DEM data.

\begin{figure}[ht]
    \centering
    \subfloat[mound 1][Mound 1]{
        \includegraphics[width=0.175\textwidth]{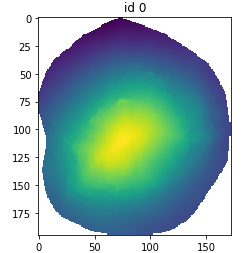}\label{fig:Mound 1}
    }
    \subfloat[mound 2][Mound 2]{
        \includegraphics[width=0.175\textwidth]{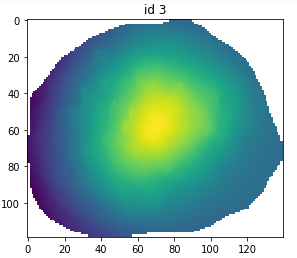}\label{fig:Mound 2}
    }
    \newline
    \subfloat[mound 3][Mound 3]{
        \includegraphics[width=0.175\textwidth]{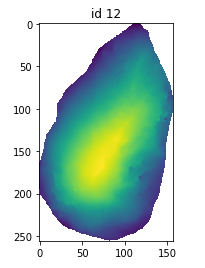}\label{fig:Mound 3}
    }
    \caption{Mounds samples from the training dataset.}
\end{figure}
We mask the labels using rasterio.mask.mask()\footnote{\url{https://rasterio.readthedocs.io/en/latest/api/rasterio.mask.html}}, which takes the dataset, the geometry of the mounds (stored in the shape file) and the noDataValue of the DEM. 
The annotations would be:
\begin{itemize}
    \item ‘0’ For non-mounds points.
    \item ‘1’ For mounds points.
\end{itemize}
We also tried treating the invalid data as an independent class but the results were worsened so we chose only those two classes.
\begin{figure}[ht]
    \centering
    \subfloat[annotated image][Annotated Image]{
        \includegraphics[width=0.3\textwidth]{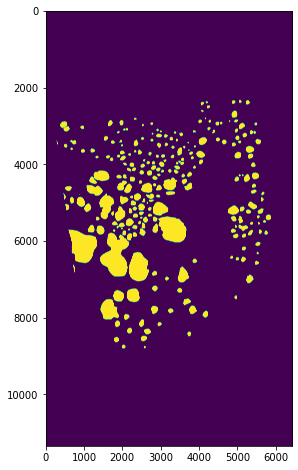}\label{fig:Annotated Image}
    }
    \caption{The annotated image (mask) is obtained by overlapping the coordination of the annotated mounds (provided in the shape file of the DEM) with the original DEM.}
\end{figure}


\section{MODEL IMPLEMENTATION}

\subsection{Methodology }
For our task, we first have to train an image segmentation network to extract the mounds from the DEM tiles. We then train the same image segmentation network with additional synthetic training data generated by a generative network, and observe if the additional augmentation data improves the accuracy of the segmentation. For the proposed technique, we need two neural network architectures, one for image segmentation and other for image generation. We chose to try two architectures for the segmentation of mounds, U-Net and FPN and later compare them. GAN was chosen for the image and label generation.

\subsection{U-NET}
U-NET \cite{RFB15a} is deep convolutional network, developed for semantic segmentation of biomedical images. It is called a "U-NET" because its architecture has a "U" shape to it. It learns to segment images in an end to end setting, meaning it gets raw input image in, and outputs a segmentation map.

Image is fed into the input layer, then the data gets propagated through the network among all the possible paths. Most operations are 3x3 convolutions, followed by a non linear activation functions (ReLU). During the 3x3 convolutions, one pixel order is lost, this allows the network to process large image in individual tiles. Next operation is max pooling, it reduces the X,Y size of the feature map, illustrated by a downward arrow. Max pooling propagates max activation from each 2x2 window to the next feature map, preserving all the important features. Sequence of convolutions and max pooling results in 'spatial contraction', which means we gradually increase the "what" in the image at the same time decrease the "where". This allows for a kind of generalization of the image structures.

  UNET also has an expansion path upwards, after the contraction. This creates a high resolution segmentation map. The expansion path has a sequence of up convolutions and concatenations, which correspond to high resolution features from the contracting path. The output we acquire is a segmentation map with two channels, one channel for foreground and one channel for the background. In our case, the foreground will be the labelled mound shapes.

We chose to use UNET for our segmentation architecture because it is observed that they outperform other segmentation architectures when the dataset size is low, i.e. few training images. Moreover, they are faster than other networks and result in faster convergence too. They also work well when two images of the same class are touching each other. The challenges for our dataset were fuzzy borders, low contrast and lack of training images.

\begin{figure}[htb]
    \centering
    \subfloat[U-NET Architecture][U-NET Architecture]{
        \includegraphics[width=0.45\textwidth]{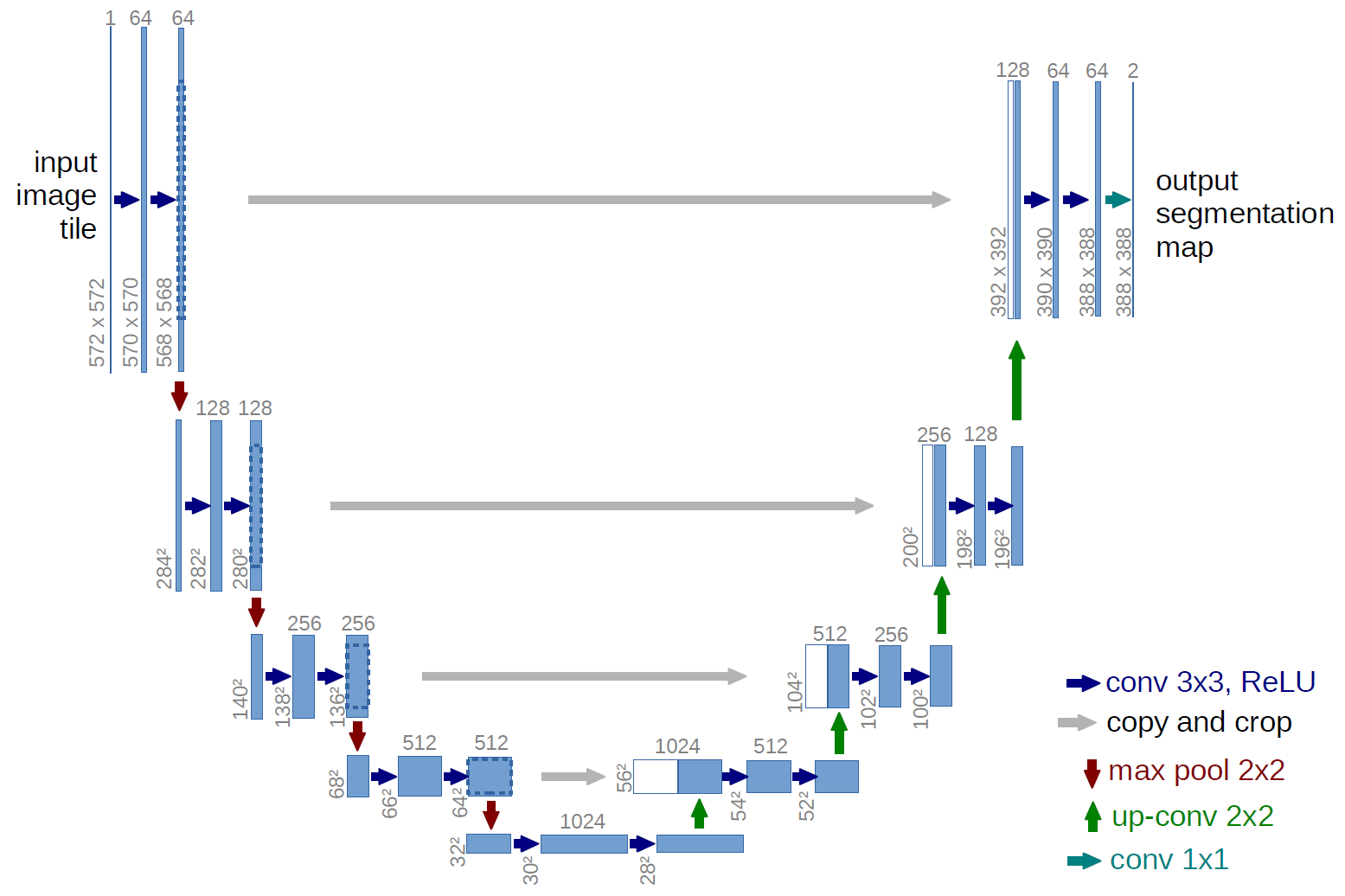}\label{fig:U-NET architecture}
    }
    \caption{U-NET architecture (https://lmb.informatik.uni-freiburg.de/people/ronneber/u-net/)}
\end{figure}

\subsection{FPN}
FPN \cite{fpn} stands for Feature Pyramid Network. It is a feature extraction neural network, commonly used for object detection. In image segmentation tasks, sometimes a model fails to detect objects that are too small or large compared to the training data sizes. FPN was introduced to tackle the problem of detecting objects in different scales.

FPN has two pathways, a bottom up and a top down pathway. The bottom up pathway uses CNNs for feature extraction. As we keep going up, the spatial resolution of the image decreases, i.e. only high level structures are preserved, that is why the semantic value or the information increases and the resolution decreases as we go up in the bottom up path. The bottom up path makes use of the ResNext architecture. Each layer we move up, the spatial dimension of feature maps is reduced by half.

After the bottom up path, FPNs have a top down path that takes the low resolution, high semantic feature maps and passes it down, increasing the image resolution. This results in a semantically rich layer at the end of the top down path. The reconstructed layers might contain high semantic information but the location of the objects is poor. This happens due to the upsampling and downsampling applied in the bottom up and top down path. Therefore, we add lateral connections between each of the pathways, this will help the detector predict locations of the objects more precisely. These lateral connections between the two paths are often referred to as skip connections.

For segmentation, a 2D sliding window is passed over the feature maps and generates a object segment, this is done for all the feature maps. Finally we combine all the object segments at different scale and create our final mask prediction.

\begin{figure}[htb]
    \centering
    \subfloat[FPN Architecture][FPN architecture]{
        \includegraphics[width=0.5\textwidth]{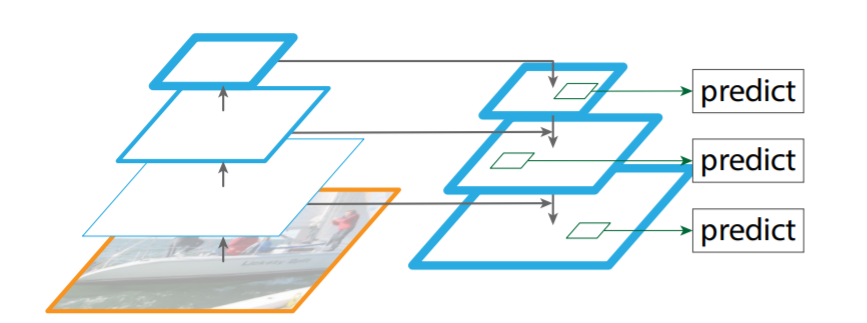}\label{fig:FPN architecture}
    }
    \caption{FPN architecture (FPN architecture)}
\end{figure}

\subsection{Pix2Pix}
GAN \cite{gan} stands for Generative Adversarial network. They are unsupervised neural networks used for image generation and manipulation. GANs can capture and copy variation in a dataset, and use it to produce new synthetic images. 

Pix2Pix is a version of conditional GANs or cGANs. This is different from normal GAN because they provide control over what image is being generated i.e. they allow us to generate an image of a given class. We need to use cGANs because we need to generate additional image data and label data as well. cGANs are generally used for image to image translation.

The architecture consists of two models, a generator network and a discriminator network. Both of these models act in an adversarial fashion, i.e against each other. The generator given an input image and is tasked with generating a translated version of the image, where as the discriminator is given an input image and real or generated paired image and is tasked with determining if the paired image is real or generated. The generator learns from distribution of classes and is supposed to fool the discriminator into thinking that the image is produced was real. As the generator gets better, discriminator also gets better. The whole network converges when discriminator can no longer differentiate between real images and the ones generated by the generator network. We also use a U-NET as a generator in our Pix2Pix model.  
  
A cGAN can be defined by the mathematical equation

\begin{equation}
\begin{aligned}
    min_{G} max_{D} V(D,G) =  E_{x~ Pdata(x)} [log D(x|y)] + E_{z~p_z}(z) \\  [log(1-D(G(z|y)))]
\end{aligned}
\end{equation}  
\hspace{0em} {$where$},
\begin{itemize}

\item G = Generator
\item D = Discriminator
\item x = sample from real data
\item z = sample from generator
\item y = Auxiliary information
\item pdata(x) = distribution of real data
\item D(x) = Discriminator network
\item G(z) = Generator network
\end{itemize}

\begin{figure}[htp]
    \centering
        \includegraphics[width=0.5\textwidth]{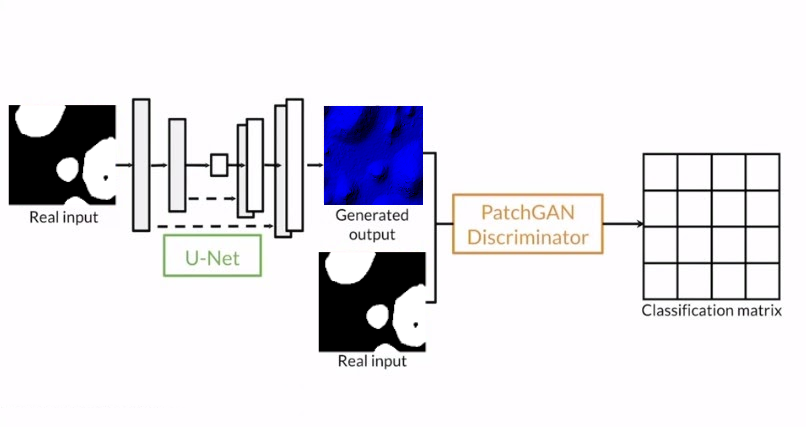}\label{fig:U-StyleGAN3 architecture}
    
    \caption{Pix2Pix architecture}
\end{figure}





\subsection{Training}

For training the U-Net and the FPN we used 2728 images in total. Of which, 1632 were used for the training dataset and 546 for the testing and validation set each. These images were obtained after the feature engineering step where we combine channels of original DEM, slope, hillshade. The dimension of the images is 224 x 192 x 3. All the images have 3 channels.

For the FPN network we used a pretrained model with se\textunderscore resnext\textunderscore 50 \cite{resnext} encoder weights pretrained on ImageNet dataset with 25M parameters from the segmentation-models-pytorch \cite{smp} library. And for the U-Net we used a pretrained resnet34\cite{resnet} encoder with weights trained on ImageNet dataset. We trained both of these models on our dataset.   

We treated our segmentation problem as a binary classification problem since the mask we predict and the ground truth mask are both binary. The predicted mask had values ranging from 0 to 1. So  we chose sigmoid as our activation function and 0.4 as the threshold. This value was chosen because it provided the best results compared to other values. For the training segmentation masks, we used images of 224 x 192 where white pixels represent mound shape and black pixels represent the background. 

The results were obtained after training the network for 10 epochs, with a batch size of 8. The learning rate used was 1e-4. 


For the GAN, we used pix2pix as our base architecture, to produce the synthesized images. The training set consisted 512 images. Out of which the test set contained 112 samples and the training set consisted of 400 samples. All images were 256 x 256 x 3 with three channels corresponding to DEM, slope and hillshade feature respectively. The model takes a lot of time to converge, even with the small dataset. The generator model had over 41M tunable parameters.  The model was trained for 120 epochs and generated 128 synthetic images with labels as the output. The generation was done on the labels of the test dataset.  

The output of the GAN and ground truth had a Bhattacharyya distance of 0.4610 and a mutual information score of 0.32 but the predicted image visually appeared very similar to the real input image. 


\begin{figure}[ht]
    \centering
    \subfloat{
        \vspace*{-3cm}\includegraphics[width=0.45\textwidth]{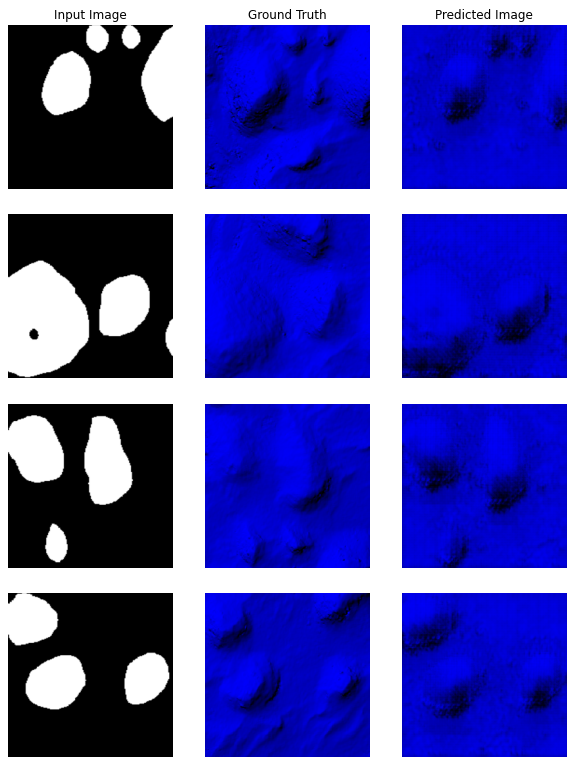}\label{}
    }
    \caption{GAN generated output results}
\end{figure}

The images had very natural looking textures and shadows. Some outputs had a grainy appearance or presence of some artifacts such as repeating patterns. It was observed that the artifacts and artificial looking patterns started slowly disappearing as the training went on. 
In some cases, it was virtually impossible to distinguish if the prediction was a real image or fake image.

\vspace{5em}


    
    

\clearpage
\newpage
\section{EVALUATION}
\subsection{Results}
First of all, the evaluation of the generated synthetic data by GAN, was based on the same strategies used by the authors in the pix2pix article \cite{pix2pix}

\begin{figure}[!ht]
    \centering
    
        \includegraphics[width=0.35\textwidth]{"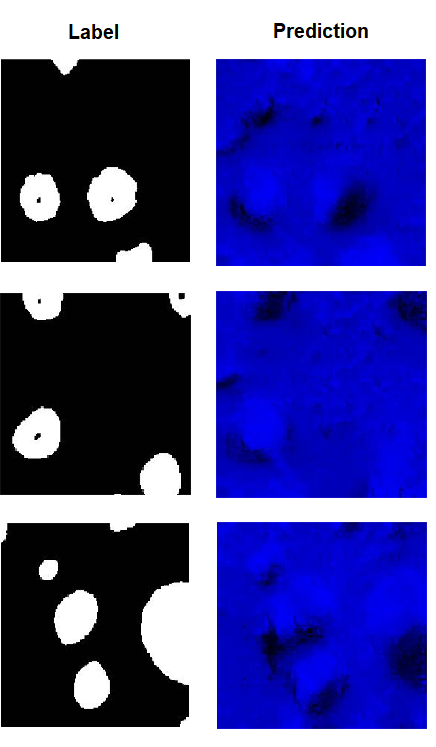"}
    
    \caption{Examples of generated augmented data}
    \label{fig:GAD}
\end{figure}

- The first strategy we used imply using human scoring. We
iterated over the generated mounds images and interpret the quality of the generated data based on human perception. Fig.\ref{fig:GAD} shows a sample of 3 images generated by our GAN model.

- The second strategy consist of rating the performances of 
a semantic segmentation network model while segmenting the real data and while segmenting the synthetic generated data. The model used in our case is U-Net pre-trained model.

In this context, we will investigate whether the U-net model and the FPN are able to segment data, and whether its performance improves by applying GAN for  data augmentation.

The segmentation quality have been evaluated using Intersection over Union (IoU)  with Dice Loss function. This loss primarily measures the overlap between samples. the measure could range from 0 to 1 where the coefficient 1 indicates perfect and complete overlap.

FPN model and U-net model trained with the real data, showed good some results.

\begin{figure}[ht]
    \centering
    
        \includegraphics[width=0.75\textwidth]{"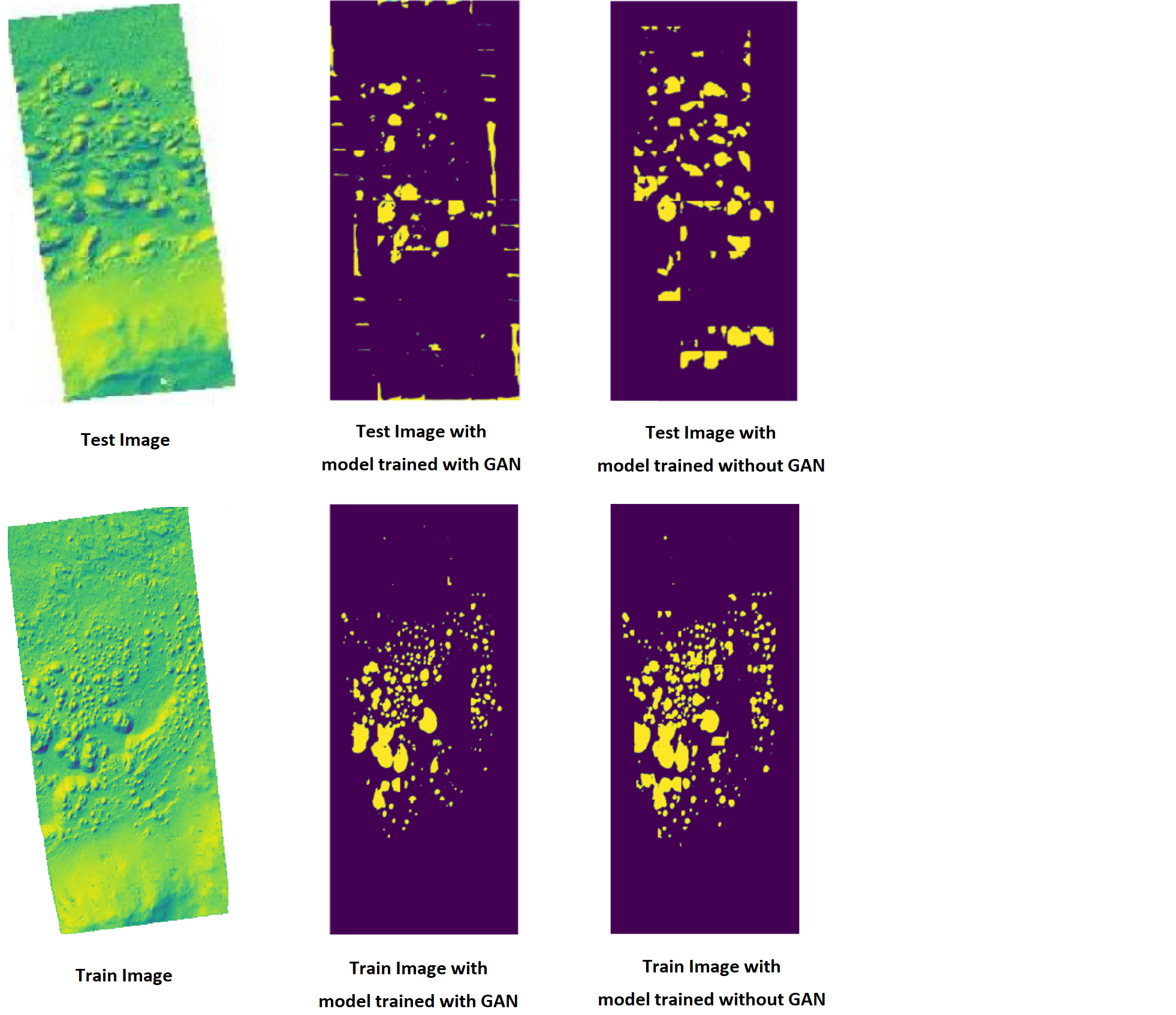"}
    
    \caption{Comparison on predicted mask}
    \label{fig:GAN}
\end{figure}

However, in the case of training these two architectures with both real and synthetic data set,  showed significantly worse quality of segmentation compared to the use of real data only As shown in Fig.\ref{fig:GAN}.

To compare both U-Net and FPN models performance, we tested the trained models with augmented data and real data using F1 score as a metric.
Results are shown in the table 1.

\begin{table}[htbp]

\begin{tabular}{
| p{\dimexpr 0.20\linewidth-2\tabcolsep} |
p{\dimexpr0.20\linewidth-2\tabcolsep}  |
p{\dimexpr0.20\linewidth-2\tabcolsep}  |
p{\dimexpr0.20\linewidth-2\tabcolsep}  |
p{\dimexpr0.20\linewidth-2\tabcolsep}  |} \hline

  &U-Net without GAN&U-Net with GAN&FPN without GAN& FPN with GAN
\\
\hline

 Accuracy &0.95&0.97&0.97&0.94
\\
\hline

 Precision &0.77&0.80&0.75&0.78
\\
\hline

 Recall &0.77&0.68&0.83&0.73
\\
\hline

FDR &0.23&0.20&0.25&0.23
\\
\hline

FOR &0.02&0.03&0.02&0.03
\\
\hline

F1-Score &0.77&0.74&0.79&0.75
\\
\hline
IOU-Score &0.81&0.81&0.84&0.80
\\
\hline

\end{tabular}
\vspace{1em}
\caption{Test metrics results}\label{tab1}

\end{table}

\subsection{Discussion}
We note that, The geographical morphologies are easily learned by GANs, and they are successfully able to reproduce very natural looking patterns features. Making them an excellent way to generate more data close to real mounds images.
Besides, the usage of two GAN models of pix2pix architecture to generate the synthetic images gave us more variance in our dataset, as both the models produced slightly different images. The final GAN synthesized additional data comprised of both these models’ output.

However, despite not having a testing mask, we can realize by observing the testing results that the segmentation with real and synthesized data sets achieved better segmentation quality than segmentation on the real data set. 
In fact, GAN confuses the network, we can discern for exemple that the model trained with GAN predicts several mounds as a line shaped especially for the case of no data areas which not correct in reality.

Finally, although we got better precision, the recall gets worse drastically which illustrate the fact that the model doesn't know how to recognize the mounds any more, in other words GAN ruin what the U-net and FPN has learned from the actual tiles.
\\

\section{Conclusion}
In this study, we investigate whether artificially generated images of the mounds on Mars can improve the performance of a segmentation model.
In this regard, we proposed and developed an FPN based network a U-Net based network for solving mound's segmentation problem. Hence, there is still the issue of training our segmentation model itself based on a limited data set. To facilitate this, we performed some transformations to the images in the original data set. Then, we leverage an existing conditional generative adversarial network (cGAN) model pretrained on large-scale data set to generate realistic synthesized data set, following the concept of transfer learning. For this purpose, we used pix2pix pre-trained to translate labels to facades.

Afterword, we compared the segmentation quality of the U-Net trained model with and without augmentation.

The results showed that models trained with real data sets achieved better segmentation quality than the one trained on both the real and synthesized data set.

In the end, we found out that adding GAN augmented data doesn't actually help in segmentation. As a matter of fact, The performance is significantly better without using GAN images, hence it will be better to rely on the real data to make better predictions.

\newpage
\vspace{10em}
\printbibliography{}

@Book{Yuval,
    Author  =       {Yuval Noah Harari},
    Title   =       {Sapiens: A Brief History of Humankind},
    Publisher   =   {Harper},
    Year    =       {2015},
    Address =       {195 Broadway New York, NY 10007 USA}
}

@Book{ForsythPonce,
  author = {David A. Forsyth and Jean Ponce},
  isbn = {978-0-273-76414-4},
  pages = {1-91},
  publisher = {Pitman},
  title = {Computer Vision - A Modern Approach, Second Edition.},
  year = {2012},
  address = {Hoboken, New Jersey}
}

@Incollection{YingTan,
title = {Chapter 11 - Applications},
editor = {Ying Tan},
booktitle = {Gpu-Based Parallel Implementation of Swarm Intelligence Algorithms},
publisher = {Morgan Kaufmann},
pages = {167-177},
year = {2016},
isbn = {978-0-12-809362-7},
doi = {https://doi.org/10.1016/B978-0-12-809362-7.50011-X},
author = {Ying Tan},
address = {San Francisco, CA}
}

@article{DassDevi,
    author = {Rajeshwar Dass and Swapna Devi},
    title = {Image Segmentation Techniques},
    year = {2012},
    journal = { International Journal of Electronics \& Communication
               Technology},
    volume = {3},
    issue = {1},
    issn = { 2230-7109 (Online)}
}

@article{PhamXuPrince,
author = {Dzung L. Pham and Chenyang Xu and Jerry L. Prince},
title = {Current Methods in Medical Image Segmentation},
journal = {Annual Review of Biomedical Engineering},
volume = {2},
number = {1},
pages = {315-337},
year = {2000},
doi = {10.1146/annurev.bioeng.2.1.315},
    note ={PMID: 11701515},

URL = { 
        https://doi.org/10.1146/annurev.bioeng.2.1.315
    
}
}

@article{ZengLiMengYangLiu,
title = {Improving histogram-based image contrast enhancement using gray-level information histogram with application to X-ray images},
journal = {Optik},
volume = {123},
number = {6},
pages = {511-520},
year = {2012},
issn = {0030-4026},
doi = {https://doi.org/10.1016/j.ijleo.2011.05.017},
author = {Ming Zeng and Youfu Li and Qinghao Meng and Ting Yang and Jian Liu},
}

@article{SaifHammadAlqubati,
    title = {Gradient Based Image Edge Detection},
    journal = {IACSIT International Journal of Engineering and Technology},
    volume = {8},
    number = {3},
    year = {2016},
    author = {Jamil A. M. Saif and Mahgoub H. Hammad and Ibrahim A. A.
              Alqubati},
    doi = { 10.7763/IJET.2016.V8.876}
}

@article{ElazizBhattacharyyaLu,
title = {Swarm selection method for multilevel thresholding image segmentation},
journal = {Expert Systems with Applications},
volume = {138},
pages = {112818},
year = {2019},
issn = {0957-4174},
doi = {https://doi.org/10.1016/j.eswa.2019.07.035},
author = {Mohamed {Abd Elaziz} and Siddhartha Bhattacharyya and Songfeng Lu},
}

@article{HousseinEmamAli,
title = {An efficient multilevel thresholding segmentation method for thermography breast cancer imaging based on improved chimp optimization algorithm},
journal = {Expert Systems with Applications},
volume = {185},
pages = {115651},
year = {2021},
issn = {0957-4174},
doi = {https://doi.org/10.1016/j.eswa.2021.115651},
author = {Essam H. Houssein and Marwa M. Emam and Abdelmgeid A. Ali},
}

@article{ChengWang,
title = {Improved region growing method for image segmentation of three-phase materials},
journal = {Powder Technology},
volume = {368},
pages = {80-89},
year = {2020},
issn = {0032-5910},
doi = {https://doi.org/10.1016/j.powtec.2020.04.032},
author = {Zhuang Cheng and Jianfeng Wang},
}

@article{Jothiaruna,
title = {A segmentation method for disease spot images incorporating chrominance in Comprehensive Color Feature and Region Growing},
journal = {Computers and Electronics in Agriculture},
volume = {165},
pages = {104934},
year = {2019},
issn = {0168-1699},
doi = {https://doi.org/10.1016/j.compag.2019.104934},
author = {Nagaraj Jothiaruna and Joseph K. {Abraham Sundar} and Balasubramanian Karthikeyan},
}

@article{LiuSclaroff,
title = {Deformable model-guided region split and merge of image regions},
journal = {Image and Vision Computing},
volume = {22},
number = {4},
pages = {343-354},
year = {2004},
issn = {0262-8856},
doi = {https://doi.org/10.1016/j.imavis.2003.11.006},
author = {Lifeng Liu and Stan Sclaroff},
}

@article{Lachaize,
title = {Evidential split-and-merge: Application to object-based image analysis},
journal = {International Journal of Approximate Reasoning},
volume = {103},
pages = {303-319},
year = {2018},
issn = {0888-613X},
doi = {https://doi.org/10.1016/j.ijar.2018.10.008},
author = {Marie Lachaize and Sylvie {Le Hégarat-Mascle} and Emanuel Aldea and Aude Maitrot and Roger Reynaud},
}

@article{KassWT,
    title = {Snakes: Active contour models},
    journal = {nternational Journal of Computer Vision},
    volume = {1},
    pages = {321–331},
    year = {1988},
    doi = {https://doi.org/10.1007/BF00133570},
    author = {Michael Kass and Andrew Witkin and Demetri Terzopoulos}
}

@article{WangWWF,
title = {A multi-object image segmentation C–V model based on region division and gradient guide},
journal = {Journal of Visual Communication and Image Representation},
volume = {39},
pages = {100-106},
year = {2016},
issn = {1047-3203},
doi = {https://doi.org/10.1016/j.jvcir.2016.05.011},
author = {Xianghai Wang and Yu Wan and Rui Li and Jinling Wang and Lingling Fang},
}

@INPROCEEDINGS{YanCaiGaoLuo,
  author={Man Yan and Jianyong Cai and Jiexing Gao and Lili Luo},
  booktitle={2012 5th International Conference on BioMedical Engineering and Informatics}, 
  title={K-means cluster algorithm based on color image enhancement for cell segmentation}, 
  year={2012},
  volume={},
  number={},
  pages={295-299},
  doi={10.1109/BMEI.2012.6513157}
}

@ONLINE {BigdataAILab,
    author = {BigdataAILab},
    title  = {What is Semantic Segmentation, Instance Segmentation, Panoramic
              segmentation?},
    month  = {04},
    year   = {2021},
    url    = {https://becominghuman.ai/what-is-semantic-segmentation-instance-segmentation-panoramic-segmentation-3bbb03856c12},
    Addendum = {{accessed: 18-11-2021}}
}

@misc{ren2016faster,
      title={Faster R-CNN: Towards Real-Time Object Detection with Region Proposal Networks}, 
      author={Shaoqing Ren and Kaiming He and Ross Girshick and Jian Sun},
      year={2016},
      eprint={1506.01497},
      archivePrefix={arXiv},
      primaryClass={cs.CV}
}

@misc{he2018mask,
      title={Mask R-CNN}, 
      author={Kaiming He and Georgia Gkioxari and Piotr Dollár and Ross Girshick},
      year={2018},
      eprint={1703.06870},
      archivePrefix={arXiv},
      primaryClass={cs.CV}
}

@misc{visin2016reseg,
      title={ReSeg: A Recurrent Neural Network-based Model for Semantic Segmentation}, 
      author={Francesco Visin and Marco Ciccone and Adriana Romero and Kyle Kastner and Kyunghyun Cho and Yoshua Bengio and Matteo Matteucci and Aaron Courville},
      year={2016},
      eprint={1511.07053},
      archivePrefix={arXiv},
      primaryClass={cs.CV}
}

@misc{luc2016semantic,
      title={Semantic Segmentation using Adversarial Networks}, 
      author={Pauline Luc and Camille Couprie and Soumith Chintala and Jakob Verbeek},
      year={2016},
      eprint={1611.08408},
      archivePrefix={arXiv},
      primaryClass={cs.CV}
}

@INPROCEEDINGS{8237868,
  author={Souly, Nasim and Spampinato, Concetto and Shah, Mubarak},
  booktitle={2017 IEEE International Conference on Computer Vision (ICCV)}, 
  title={Semi Supervised Semantic Segmentation Using Generative Adversarial Network}, 
  year={2017},
  volume={},
  number={},
  pages={5689-5697},
  doi={10.1109/ICCV.2017.606}
}

@misc{hung2018adversarial,
      title={Adversarial Learning for Semi-Supervised Semantic Segmentation}, 
      author={Wei-Chih Hung and Yi-Hsuan Tsai and Yan-Ting Liou and Yen-Yu Lin and Ming-Hsuan Yang},
      year={2018},
      eprint={1802.07934},
      archivePrefix={arXiv},
      primaryClass={cs.CV}
}

@online{ganzoo,
    url = {https://github.com/hindupuravinash/the-gan-zoo},
    addendum = {accessed: 18-11-2021}
}

@misc{ronneberger2015unet,
      title={U-Net: Convolutional Networks for Biomedical Image Segmentation}, 
      author={Olaf Ronneberger and Philipp Fischer and Thomas Brox},
      year={2015},
      eprint={1505.04597},
      archivePrefix={arXiv},
      primaryClass={cs.CV}
}

@misc{milletari2016vnet,
      title={V-Net: Fully Convolutional Neural Networks for Volumetric Medical Image Segmentation}, 
      author={Fausto Milletari and Nassir Navab and Seyed-Ahmad Ahmadi},
      year={2016},
      eprint={1606.04797},
      archivePrefix={arXiv},
      primaryClass={cs.CV}
}

@article{PaneruJeelani,
title = {Computer vision applications in construction: Current state,
         opportunities \& challenges},
journal = {Automation in Construction},
volume = {132},
pages = {103940},
year = {2021},
issn = {0926-5805},
doi = {https://doi.org/10.1016/j.autcon.2021.103940},
author = {Suman Paneru and Idris Jeelani},
}

@misc{RMC,
      title={Unsupervised Representation Learning with Deep Convolutional Generative Adversarial Networks}, 
      author={Alec Radford and Luke Metz and Soumith Chintala},
      year={2016},
      eprint={1511.06434},
      archivePrefix={arXiv},
      primaryClass={cs.LG}
}

@INPROCEEDINGS{SpickCW,
  author = {Ryan J. Spick and Peter Cowling and James Alfred Walker},
  booktitle = {2019 IEEE Conference on Games (CoG)}, 
  title = {Procedural Generation using Spatial GANs for Region-Specific Learning of Elevation Data}, 
  year={2019},
  pages={1-8},
  doi={10.1109/CIG.2019.8848120}
}

@misc{JetchevBV,
      title={Texture Synthesis with Spatial Generative Adversarial Networks}, 
      author={Nikolay Jetchev and Urs Bergmann and Roland Vollgraf},
      year={2017},
      eprint={1611.08207},
      archivePrefix={arXiv},
      primaryClass={cs.CV}
}

@misc{bowles2018gan,
      title={GAN Augmentation: Augmenting Training Data using Generative Adversarial Networks}, 
      author={Christopher Bowles and Liang Chen and Ricardo Guerrero and Paul Bentley and Roger Gunn and Alexander Hammers and David Alexander Dickie and Maria Valdés Hernández and Joanna Wardlaw and Daniel Rueckert},
      year={2018},
      eprint={1810.10863},
      archivePrefix={arXiv},
      primaryClass={cs.CV}
}

@misc{KarrasAJ,
      title={Progressive Growing of GANs for Improved Quality, Stability, and Variation}, 
      author={Tero Karras and Timo Aila and Samuli Laine and Jaakko Lehtinen},
      year={2018},
      eprint={1710.10196},
      archivePrefix={arXiv},
      primaryClass={cs.NE}
}

@InProceedings{RFB15a,
  author       = "O. Ronneberger and P.Fischer and T. Brox",
  title        = "U-Net: Convolutional Networks for Biomedical Image Segmentation",
  booktitle    = "Medical Image Computing and Computer-Assisted Intervention (MICCAI)",
  series       = "LNCS",
  volume       = "9351",
  pages        = "234--241",
  year         = "2015",
  publisher    = "Springer",
  note         = "(available on arXiv:1505.04597 [cs.CV])",
  url          = "http://lmb.informatik.uni-freiburg.de/Publications/2015/RFB15a"
}

@misc{gan,
      title={Generative Adversarial Networks}, 
      author={Ian J. Goodfellow and Jean Pouget-Abadie and Mehdi Mirza and Bing Xu and David Warde-Farley and Sherjil Ozair and Aaron Courville and Yoshua Bengio},
      year={2014},
      eprint={1406.2661},
      archivePrefix={arXiv},
      primaryClass={stat.ML}
}

@misc{smp,
  Author = {Pavel Yakubovskiy},
  Title = {Segmentation Models Pytorch},
  Year = {2020},
  Publisher = {GitHub},
  Journal = {GitHub repository},
  Howpublished = {\url{https://github.com/qubvel/segmentation_models.pytorch}}
}

@article{resnext,
  author    = {Saining Xie and
               Ross B. Girshick and
               Piotr Doll{\'{a}}r and
               Zhuowen Tu and
               Kaiming He},
  title     = {Aggregated Residual Transformations for Deep Neural Networks},
  journal   = {CoRR},
  volume    = {abs/1611.05431},
  year      = {2016},
  url       = {http://arxiv.org/abs/1611.05431},
  eprinttype = {arXiv},
  eprint    = {1611.05431},
  bibsource = {dblp computer science bibliography, https://dblp.org}
}

@article{FPN,
  author    = {Tsung{-}Yi Lin and
               Piotr Doll{\'{a}}r and
               Ross B. Girshick and
               Kaiming He and
               Bharath Hariharan and
               Serge J. Belongie},
  title     = {Feature Pyramid Networks for Object Detection},
  journal   = {CoRR},
  volume    = {abs/1612.03144},
  year      = {2016},
  url       = {http://arxiv.org/abs/1612.03144},
  eprinttype = {arXiv},
  eprint    = {1612.03144},
  bibsource = {dblp computer science bibliography, https://dblp.org}
}

@article{resnet,
  author    = {Kaiming He and
               Xiangyu Zhang and
               Shaoqing Ren and
               Jian Sun},
  title     = {Deep Residual Learning for Image Recognition},
  journal   = {CoRR},
  volume    = {abs/1512.03385},
  year      = {2015},
  url       = {http://arxiv.org/abs/1512.03385},
  eprinttype = {arXiv},
  eprint    = {1512.03385},
  bibsource = {dblp computer science bibliography, https://dblp.org}
}

@article{pix2pix,
  author    = {Phillip Isola and Jun-Yan Zhu and Tinghui Zhou and Alexei A Efros},
  title     = {Image-toImage Translation with Conditional Adversarial Networks},
  journal   = {In Computer Vision
and Pattern Recognition (CVPR)},
  year      = {2017},
  bibsource = {2017 IEEE Conference on}
}
\appendix
\end{document}